\documentclass[lettersize,journal]{IEEEtran}
\usepackage{amsmath,amsfonts}
\usepackage{array}
\usepackage{textcomp}
\usepackage{stfloats}
\usepackage[hidelinks]{hyperref}
\usepackage{verbatim}
\usepackage{graphicx}
\def\BibTeX{{\rm B\kern-.05em{\sc i\kern-.025em b}\kern-.08em
    T\kern-.1667em\lower.7ex\hbox{E}\kern-.125emX}}

\usepackage{balance}
\usepackage{subcaption}
\usepackage{cleveref}
\usepackage{tikz}
\usepackage{booktabs} % For professional looking tables
\usepackage{multirow}

\usepackage{tabularx}
\usepackage{multicol}

\usepackage{algorithm}
\usepackage{algpseudocode}
\usepackage{mathtools}

\usetikzlibrary{intersections}
\usetikzlibrary{shapes,arrows,positioning}
\usetikzlibrary{calc,arrows.meta}
\usepackage{xparse}
\NewDocumentCommand\Cycle{O{} m m m O{} m}{
  \draw[#1](#2.{#3+asin(#6/(#4*1.41))}) arc (180+#3-45:180+#3-45-270:#6/2) #5;
} 
\tikzstyle{block} = [rectangle, draw]
\tikzstyle{input} = [coordinate]
\tikzstyle{output} = [coordinate]
\tikzstyle{pinstyle} = [pin edge={to-,thin,black}]

\usepackage{enumitem}

\usepackage[normalem]{ulem}

\usepackage{tcolorbox}

\definecolor{boxBorder}{RGB}{160, 50, 50}      % Dark red border
\definecolor{boxTitleBg}{RGB}{218, 157, 157}   % Pinkish title background
\definecolor{boxBodyBg}{RGB}{248, 235, 235}    % Very light pink body

\newtcolorbox{promptbox}[1]{
    colframe=boxBorder,        % Border color
    colback=boxBodyBg,         % Body background color
    colbacktitle=boxTitleBg,   % Title background color
    coltitle=black,            % Title text color
    title={\textbf{#1}},       % The title text (bold)
    fonttitle=\bfseries,       % Font style for title
    sharp corners,             % Sharp corners like the image
    boxrule=1.5pt,             % Thicker border line
    titlerule=0.5pt,           % Thin line between title and body
    left=5pt, right=5pt, top=5pt, bottom=5pt % Padding
}
\begin{document}
\title{Nemobot Games: Crafting Strategic AI Gaming Agents for Interactive Learning with Large Language Models}
\author{Chee Wei Tan, Yuchen Wang, Shangxin Guo
%\thanks{Correspondence author: C. W. Tan, Nanyang Technological University, Singapore, Nanyang Ave., Singapore (e-mail: cheewei.tan@ntu.edu.sg).}
\thanks{C. W. Tan and Y. Wang are with Nanyang Technological University, Singapore, Nanyang Ave., Singapore (e-mail: cheewei.tan@ntu.edu.sg).}
\thanks{S. Guo is with Nautilus Software Technologies Limited, Hong Kong.}
}

%\markboth{Journal of \LaTeX\ Class Files,~Vol.~18, No.~9, September~2020}%
%{How to Use the IEEEtran \LaTeX \ Templates}

\maketitle

\begin{abstract}
This paper introduces a new paradigm for AI game programming, leveraging large language models (LLMs) to extend and operationalize Claude Shannon's taxonomy of game-playing machines. Central to this paradigm is {\it Nemobot Games}, an interactive agentic engineering environment that enables users to create, customize, and deploy LLM-powered game agents while actively engaging with AI-driven strategies. The LLM-based chatbot, integrated within Nemobot Games, demonstrates its capabilities across four distinct classes of games. For {\it dictionary-based games}, it compresses state-action mappings into efficient, generalized models for rapid adaptability. In {\it rigorously solvable games}, it employs mathematical reasoning to compute optimal strategies and generates human-readable explanations for its decisions. For {\it heuristic-based games}, it synthesizes strategies by combining insights from classical minimax algorithms (see, e.g., \cite{shannon1950chess}) with crowd-sourced data. Finally, in {\it learning-based games}, it utilizes reinforcement learning with human feedback and self-critique to iteratively refine strategies through trial-and-error and imitation learning. {\it Nemobot} amplifies this framework by offering a programmable environment where users can experiment with tool-augmented generation and fine-tuning of strategic game agents. From strategic games to role-playing games, Nemobot Games demonstrates how AI agents can achieve a form of self-programming by integrating crowdsourced learning and human creativity to iteratively refine their own logic. This represents a step toward the long-term goal of self-programming AI.
\end{abstract}


\section{Introduction}
\IEEEPARstart{C}{laude} Shannon, a pioneer in artificial intelligence, posed a thought-provoking question in his seminal work \cite{shannon1953computers}: {\it ``Can we program a digital computer so that (eventually) 99 percent of the orders it follows are written by the computer itself, rather than the few percent in current programs?"} This question encapsulates the vision of self-programming systems, a concept that remains a core challenge in AI research. Similarly, Marvin Minsky argued in \cite{minsky1967programming} that programming's structured rules are not limitations but tools for refining and clarifying poorly understood ideas, akin to how artists and composers use formal constraints to express creativity. Shannon's vision of computers autonomously generating their own instructions has become increasingly relevant in agentic engineering, where large language models (LLMs) now produce vast amounts of code, driving self-evolving AI systems and interactive applications such as games.

\begin{figure}[t]
  \centering
\includegraphics[width=0.5\textwidth]{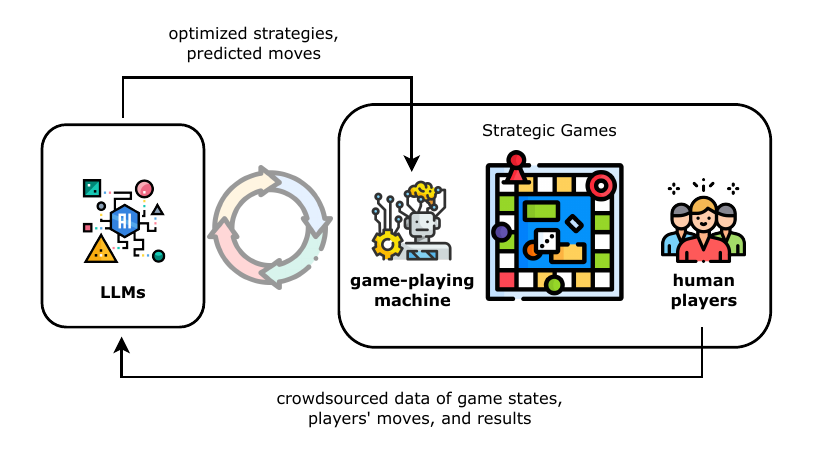}
  \caption{Crowdsourcing and strategy optimization in game-playing AI. The LLM generates optimized strategies for game-playing agents, while game states and results from interactions with human players are fed back to train the LLM, creating a self-reinforcing cycle of improvement through crowdsourcing.}\label{crowd_sourcing_loop}
  % \vspace{-2mm}
\end{figure}

\begin{figure*}[tbp]
    \centering
    \includegraphics[width=0.8\linewidth]{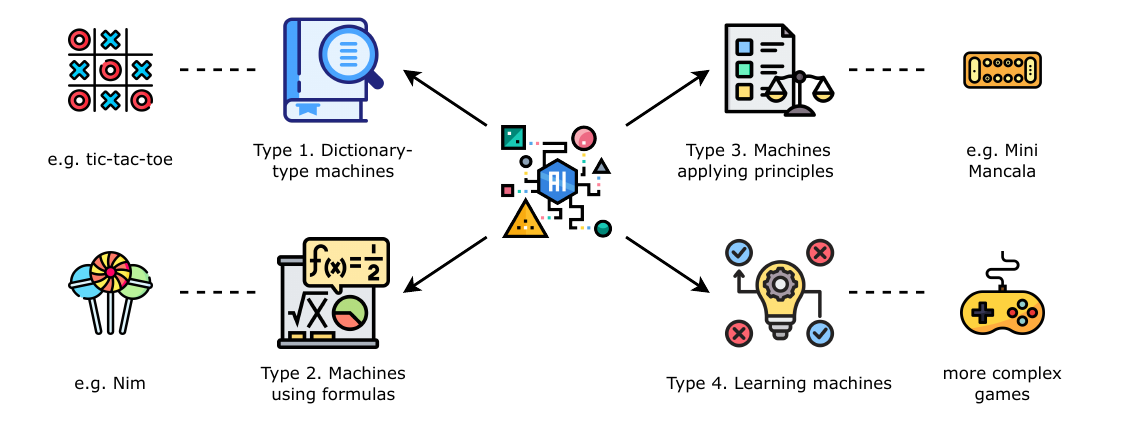}
    \caption{Shannon's four types of game-playing machines and their typical examples of applications.}
    \label{four_types}
\end{figure*}

Recent surveys, such as \cite{hu2024survey,huang2024agent}, explore advances in strategic AI game agents to enhance adaptability and reasoning using LLMs. Research into specific applications such as combining language models with strategic reasoning in the game of Diplomacy \cite{bakhtin2022humanlevel} and using LLMs as testers to measure game difficulty \cite{xiao2023llm} demonstrates the versatility of LLMs in imperfect-information games. Approaches like ChessGPT \cite{feng2024chessgpt} and AlphaStar \cite{deepmind2019alphastar} bridge policy learning with language modeling, emphasizing the importance of synergizing reinforcement learning and natural language processing. The development of systems such as Suspicion-Agent \cite{guo2024suspicionagent}, which integrates theory-of-mind-aware reasoning with OpenAI's GPT-4, further extends the horizon of multi-agent interactions in imperfect information settings.

Game boards serve as microworlds: discrete, observable world states that provide a structured, controllable testbed for studying {\it representation learning of world models} in LLMs. When trained on game transcripts and rules, LLMs can learn implicit representations that encode not only legal moves but also temporal dependencies, hidden states, and probabilistic opponent strategies. For instance, \cite{toshniwal2022chess} demonstrate that chess gameplay enables LLMs to track board states and predict moves, providing evidence that strategic environments can induce meaningful internal representations of game dynamics. This aligns with findings by \cite{li2023emergent}, who show that transformers trained on synthetic sequential prediction tasks spontaneously form internal world models capturing latent environmental dynamics without explicit supervision. To rigorously evaluate the fidelity of these learned representations, \cite{vafa2024evaluating} propose systematic methods for probing whether generative models maintain consistent beliefs about unobserved variables and adhere to transition rules. Collectively, these studies suggest that gameplay can drive \textit{representational learning} of world models in LLMs though these models often remain approximate and lack explicit grounding.

To refine and leverage these representations, {\it AI game programming} provides a powerful paradigm for structured world-model learning. By designing game simulators and programmed interfaces, we can create environments where an LLM agent must learn explicit, verifiable representations to succeed. For example, an agentic engineering framework can be designed to interact with a game engine through a structured API, forcing the LLM to explicitly represent board states, action spaces, and opponent models to generate valid function calls. This approach uses \textit{programmable feedback loops} to ground abstract internal representations: the LLM’s predictions are immediately validated or corrected by the game engine, enabling error-driven refinement of its world model. Through repeated interaction with these programmed interfaces, an AI agent can learn more consistent, accurate, and generalizable representations, effectively turning gameplay into a form of \textit{self-supervised representation learning}. 

These programmable approaches result in world models that are not only more interpretable but also more actionable for downstream strategic reasoning, thereby narrowing the gap between emergent internal representations and externally verifiable game semantics. This downstream strategic reasoning can be further enhanced through established methods for improving language model reasoning: self-consistency \cite{wang2022selfconsistency} refines decision quality via structured reasoning chains, while frameworks like ReAct \cite{yao2022react} and Graph of Thoughts \cite{besta2023graphoftoughts} provide formal mechanisms for reasoning and planning in complex environments. Moreover, work on retrieval-augmented strategic play \cite{wu2024spring} demonstrates how language models can dynamically incorporate external knowledge to guide exploration and improve gameplay by using LLM-based agents.

Recent advances in AI-assisted programming are transforming the development of sophisticated game AI agents. Foundation models like OpenAI's GPT harness advanced NLP and LLM capabilities to interpret human instructions and generate functional code \cite{chen2021evaluating}. This has enabled tools such as GitHub Copilot \cite{friedman2021introducing_url}, Copilot for Xcode \cite{tancopilot,camp_testing}, OpenAI Codex \cite{codex_url}, and Claude Code \cite{claude_url} to support programmers in rapidly prototyping, debugging, and optimizing agent logic, thus accelerating the design of reasoning loops, tool integration and strategic modules for game AI. The evolving multimodal capabilities of publicly-available LLM systems like OpenAI's GPT \cite{achiam2023gpt4} and Google's Gemini \cite{team2023gemini} further extend this potential, reinforcing the synergy between AI-assisted programming, strategic reasoning, and interactive gameplay. Recent frameworks like CAMP \cite{camp, camp_testing} with context-based RAG methods demonstrate how local-cloud architectures and retrieval mechanisms can further enhance the reliability and context-awareness of AI-assisted programming and software testing. These developments allow us to revisit Shannon's question through a new programming paradigm of agentic engineering, where learned representations are programmatically leveraged--rather than relying on a full world model--to create hybrid, human-AI solutions.

A natural question arises: if modern LLMs can already play games such as tic-tac-toe or Nim when prompted directly, what is the value of a dedicated programming framework? The distinction lies in the difference between \textit{intelligence} and \textit{programming}. Prompting an LLM to play a game yields non-deterministic, often non-reproducible behavior that is opaque to the user and difficult to debug or improve systematically. Shannon's taxonomy~\cite{shannon1953computers,Shannon1955GamePlaying} was developed when game-playing intelligence was largely deterministic and algorithmically grounded; however, current LLM-based systems are inherently non-deterministic. What is lacking is a structured programming paradigm that harnesses LLM capabilities within a reproducible, educational, and extensible framework. \textit{Nemobot} addresses this gap by providing a programmable environment where LLM functions are treated as modular, inspectable components---analogous to subroutines---that users can design, test, refine, and share. Rather than merely orchestrating LLM API calls, Nemobot Games operationalizes Shannon's taxonomy through programmable prompt engineering, crowdsourced strategy refinement, and a tiered curriculum that scaffolds learning from simple dictionary-based games to advanced reinforcement-learning agents. This positions the contribution not as a wrapper around LLM services, but as a research framework that investigates how classical game-playing paradigms can be systematically extended with LLMs for both AI research and education. Recent work in IEEE Transactions on Games further motivates this direction: Phillips et al.~\cite{phillips2025goal} demonstrated goal-oriented NPC interactions using LLMs, while Bassanelli et al.~\cite{bassanelli2025gamidoc} highlighted the importance of principled gamification design for effective learning outcomes.

This paper revisits Shannon's taxonomy of game-playing machines \cite{shannon1953computers,Shannon1955GamePlaying} and extends it using modern LLM capabilities to introduce a programming paradigm for strategic game development. Shannon’s four categories, illustrated in Figure \ref{four_types}: 
dictionary-based machines with pre-stored solutions, mathematically rigorous systems for solvable games, heuristic-driven machines for complex scenarios and learning machines that adapt through experience remain the foundational elements of intelligent systems. LLMs enhance these categories by compressing state-action mappings for {\it dictionary-based games}, applying symbolic reasoning for {\it rigorously solvable games}, integrating minimax algorithms and crowd-sourced data for {\it heuristic-based games}, and employing reinforcement learning with human feedback for {\it learning-based games}.  Central to this exploration is {\it Nemobot Games}, a game-playing chatbot and interactive programming environment that embodies these principles. Nemobot Games empowers users to design and deploy game-playing agents, extending Shannon’s taxonomy with LLMs. Through games like tic-tac-toe, Nim, and Mancala, it showcases LLM-driven advancements in AI game programming, integrating natural language understanding and generation.

In summary, the contributions of our work are as follows:
\begin{itemize}
\item An AI game programming framework called {\it Nemobot Games} is proposed, integrating language models with Shannon's foundational concepts of game-playing machines. This framework uses LLM-based programming to enable AI game bots that interact with humans and adapt strategies. Through LLM functions like reasoning, decision generation, and neuralized memoization, the bots efficiently store and refine strategies, enabling trial-and-error learning with human feedback.

\item A systematic approach to collaborative prompt engineering through crowdsourcing is designed to enable programmable gameplay strategies and fostering creativity in AI game development with human feedback.

\item The Nemobot Games framework is evaluated in remote learning environments, highlighting its effectiveness in gamified AI education. By enabling collaborative prompt engineering through crowdsourcing, Nemobot Games fosters interactive learning and helps students grasp foundational AI concepts.
\end{itemize}

% \begin{itemize}
% \item We propose an AI game programming framework called {\it Nemobot} Games, integrating language models with Shannon's foundational concepts of game-playing machines. This framework uses LLM-based programming to enable AI game bots that interact with humans and adapt strategies. Through LLM functions like reasoning, decision generation, and neuralized memoization, the bots efficiently store and refine strategies, enabling trial-and-error learning with human feedback.

% \item We design a systematic approach to collaborative prompt engineering through crowdsourcing, enabling the dynamic creation of programmable gameplay strategies and fostering creativity in AI game development with human feedback.

% \item We evaluate the Nemobot framework in remote learning environments, highlighting its effectiveness in gamified AI education. By enabling collaborative prompt engineering through crowdsourcing, Nemobot fosters interactive learning and helps students grasp foundational AI concepts.
% \end{itemize}

The rest of this paper is organized as follows: Section \ref{sec:llmgame} proposes the integration of LLM within Shannon's taxonomy and how to structure AI game programming with collaborative prompt engineering using LLMs. Section \ref{sec:nemobot} describes our {\it Nemobot} programming framework for programmable prompt engineering for AI games. Section \ref{sec:evaluation} discusses collaborative prompt engineering, using real-world examples to align crowdsourced human feedback with strategic games and presents trial results in a practical setting. Finally, Section \ref{furtherdiscussions} explores open issues and future work, while Section \ref{sec:conclusion} summarizes the key insights and contributions of this paper.

\section{Game Playing with Large Language Models}
\label{sec:llmgame}

\subsection{Shannon's Taxonomy of AI Games with LLMs}

For {\it dictionary-based machines}, Shannon described systems that predefine every possible game situation and store corresponding moves in a ``dictionary'' or function table \cite{shannon1953computers}. Traditional implementations require large memory resources, limiting their practicality to simple games like tic-tac-toe. 
Building on this foundational concept, we propose that language models can be utilized to augment this approach by learning patterns from game data. We hypothesize that this effectively compresses the dictionary into a smaller, generalized model, allowing the system to infer moves for previously unseen positions without exhaustive storage. This approach provides new programmers a practical way to learn AI game programming by editing predefined moves in real time, gaining hands-on experience with LLM functions, deciding when to use the LLM or local memory, and understanding strategic decision-making in interactive environments.

In the category of {\it machines using rigorous mathematical formulas}, Shannon outlined systems capable of perfect gameplay in mathematically solvable games like Nim \cite{Shannon1955GamePlaying}.
% In the proposed framework, LLMs act as interactive copilots designed to explain the reasoning behind optimal moves in natural language. This design intends to make these systems valuable to the programmer, who benefits from the model's ability to articulate the logic behind optimal moves in games like Nim.
% By integrating LLMs, these games can do more than just provide solutions; they can offer real-time programming assistance by breaking down complex calculations into step-by-step insights. This bridges the gap between algorithmic execution and strategy formulation, making mathematically-proven strategies in AI games more accessible. Additionally, the idea of AI-assisted programming \cite{wong2023natural,WongTan2024} and existing LLMs' code generation capability (e.g., GitHub Copilot~\cite{friedman2021introducing}, Copilot for Xcode~\cite{tancopilot}, Google AlphaCode~\cite{li2022competition}, Amazon CodeWhisperer~\cite{amazon} and Codeium~\cite{codeium}) can be utilized to create code for the mathematical strategies behind these games.
In the proposed framework, LLMs act as interactive copilots designed to explain the reasoning behind optimal moves in natural language. This design intends to make these systems valuable to the programmer, who benefits from the model's ability to articulate the logic behind optimal moves in games like Nim. By integrating LLMs, these games can do more than just provide solutions; they can offer real-time programming assistance by breaking down complex calculations into step-by-step insights \cite{nemobot2024}. This bridges the gap between algorithmic execution and strategy formulation, making mathematically-proven strategies in AI games more accessible. Additionally, the concept of AI-assisted programming \cite{wong2023natural,WongTan2024} and existing LLM code generation capabilities can be utilized to create code for the mathematical strategies behind these games. 

To illustrate, in the 1969 {\it The Game of Euclid} \cite{coledaviegame}, players take turns removing multiples of the smaller pile from the larger, aiming to reduce one pile to zero. The game has an optimal play strategy tied to the computation of the {\it Greatest Common Divisor (GCD)}.
We demonstrate that LLMs can go beyond analyzing the game by generating code to implement the most efficient GCD algorithms, such as the classic Euclidean algorithm or optimized variants. For example, given two piles of $a$ and $b$ stones $a > b$), an LLM can generate code to iteratively compute $a \mod b$, identify the optimal $m$ for reducing $a$ by $m \times b$, and ensure efficient computation while avoiding losing configurations. The code generation capability of LLMs enables programmers to explore both the strategic elements of the game and the efficient implementation of mathematical algorithms, optimizing computational performance through context-aware code generation.

The third category, {\it machines employing general principles of play}, addresses games like chess, poker, and bridge, where no exact solutions exist but heuristic strategies guide decision-making \cite{shannon1950chess}. Language models can synthesize extensive game-related knowledge, human strategies, and expert insights, enabling the generation of context-aware strategies for various scenarios. By incorporating techniques like Monte Carlo tree search and leveraging self-play, as demonstrated by AlphaGo \cite{silver2016mastering, silver2018general}, LLMs refine strategies through training on both human and AI-generated games. Their chain-of-thought reasoning \cite{chainofthoughts, wang2022selfconsistency} allows them to break down complex problems into step-by-step solutions, suggesting heuristics for optimal paths and recommending efficient search algorithms, thereby outsourcing computations like pathfinding and strategic exploration. This streamlines decision-making to enhance game performance with efficient algorithms, and fosters interaction between LLMs and human players who can test-drive strategies \cite{WongTan2024}. In programming these strategy games, LLMs can simulate skill levels and explain decisions through self-play, enhancing efficiency in probabilistic reasoning \cite{silver2018general}.

Finally, learning machines, the pinnacle of Shannon's taxonomy, closely align with modern data-driven approaches. Shannon envisioned machines that enhance their strategies through trial-and-error learning, continuously refining their approaches based on extensive experience \cite{shannon1950chess}. Similarly, LLMs can be integrated as adaptive tools to assist in the programming and refinement of learning through trial and error \cite{barto2019reinforcement}. LLMs can guide the development of self-improving AI opponents by explaining feedback loops and training data's role in learning. They accelerate reinforcement learning, particularly Michie's Boxes algorithm in \cite{michie1968boxes, michie1973machines}, which is a stochastic, trial-and-error system that learns to avoid failures in unknown environments and is a direct intellectual ancestor of modern reinforcement learning \cite{barto2019reinforcement}. By generating moves, evaluating outcomes, and refining strategies, LLMs act as AI coaches that simulate skilled opponents, provide gameplay feedback, and emulate reward models, while their natural language insights foster explainable strategic game play.

\subsection{Programmable Prompts via LLM Functions}

Large language models have the ability to synthesize knowledge from massive (raw and unlabelled) datasets, making them powerful tools for autonomous learning and decision-making \cite{chen2021evaluating}. We can design LLM agents that do not merely follow pre-programmed rules but can generate strategies, responses, and solutions based on the context they encounter. This approach aligns with Shannon’s vision of a machine that gradually writes more of its own orders. For example, in game-playing systems, instead of being limited to predefined strategies or rigid algorithms, a query can be made to different LLMs to generate feasible strategies to improve game learning performance iteratively with minimal human input.

Programmable prompt engineering further enhances this capacity by allowing developers to implement logical response behavior using {\it LLM functions}. Rather than writing exhaustive code to cover every possible situation, prompt engineering allows the program to dynamically understand and respond to varied situations through contextually relevant instructions in real-time. This allows for a flexible, scalable approach to problem-solving where the machine ``writes" its own orders through prompt-based guidance. LLM functions thus serve to generate strategic decisions in games, adapt its behavior based on previous outcomes, and refine its strategies through feedback loops.

Hence, instead of relying solely on pre-programmed strategies, an AI game engine can autonomously generate new strategies by interpreting prompts and adjusting its play based on real-time feedback. This shift toward greater autonomy via machine learning for conversational data (see, e.g., \cite{learningtochat}) is not only a recent technical achievement but also reflects a deeper philosophical transition in how we interact with machines, thus moving from instructing them directly to guiding them in a way that allows them to ``write" their own programs via natural language generation and understanding. Michie's contemporaneous ``memo functions" or {\it memoization} in \cite{michie1968memofunctions,michie2000discovery,barto2019reinforcement} are now understood as a form of speedup learning  that stores and reuses previously-computed results, reducing redundancy and enhancing efficiency. Hence, LLM functions can be seen as a form of {\it neuralized memoization}, offering more dynamic, context-aware responses. 

Indeed, Michie's memo functions \cite{michie1968memofunctions,michie2000discovery} find a direct modern instantiation in the KV caching used by large language models, which stores previously computed attention keys and values to reduce generation complexity from quadratic to linear. Beyond this performance optimization, the concept can be expanded in three directions. First, semantic memoization would cache by embedding similarity rather than exact input match, allowing LLMs to retrieve responses for semantically equivalent queries without recomputation. Second, procedural memoization would cache not just results but successful strategies or tool-use sequences across agentic workflows, enabling agents to reuse effective plans. Third, distributed memoization would share cached knowledge across multiple agents, allowing collective speedup learning. Thus, treating computation as a learnable resource scales from symbolic caching to {\it neuralized memoization} and multi-agent learning, connecting to reinforcement learning by enabling adaptive learning where past interactions refine game strategies and improve collaborative problem-solving \cite{barto2019reinforcement}.

\subsection{Collaborative Prompt Engineering}
Effective prompt engineering is key to leveraging LLMs for specialized tasks like game-playing but often requires experimentation and expertise \cite{christiano2017deep,chen2021evaluating}. Crowd-sourcing can potentially offer a scalable solution to harness collective intelligence to create and refine optimized LLM prompts for strategic gaming tasks \cite{WebGPT,li2022nemobot}. Indeed, the 1999 chess match between Garry Kasparov and ``The Rest of the World" exemplifies large-scale crowdsourcing, where global participants combined human intuition, computer engines, and expert advice \cite{kasparov2000againstworld}. Kasparov's narrow victory after 62 moves highlighted both the potential and limitations of collective decision-making in human-machine collaboration. Crowdsourcing offers a scalable way to improve natural language understanding by leveraging diverse participant perspectives to refine responses \cite{learningtochat,li2022nemobot}. Crowdsourcing will matter in agentic engineering by providing scalable human oversight and diverse training data that autonomous agents cannot generate on their own.

In strategic game programming, crowdsourcing enhances LLM fine-tuning by improving reinforcement learning models and simplifying evaluation. It generates prompts for strategy, move explanations, opponent simulations, and diverse gameplay styles, boosting AI adaptability. This collaborative process aggregates comprehensive datasets, uncovers high-performing patterns, and fosters innovation. By integrating these insights into LLM agents (e.g., Nemobot Games in Section \ref{sec:nemobot}), it scales human creativity, enhances human-AI augmentation, and drives new programming paradigms in strategic AI game development, as illustrated in Figure \ref{crowd_sourcing_loop}.

The learning process in complex AI game playing will involve significant trial and error, but crowdsourcing can reduce the overhead by aggregating diverse feedback and human insights. Programmable prompts amplify crowdsourcing by providing a structured framework to encode, refine, and evaluate collective human expertise. Through LLM functions, these prompts enable agents to dynamically simulate strategic decision-making and problem-solving. Crowdsourcing from a diverse community exposes LLM agents to a wide range of reasoning styles and creative strategies, aligning with Michie’s {\it neuralized memoization}. This approach allows for iterative improvement, combining learned strategies with programmable modules to enhance performance over time.

\subsection{AI Game Programming Agent}
To leverage the full potential of LLM capabilities, a tool must be developed with the following characteristics:

\begin{itemize}
\item {\it Optimize both computational efficiency and strategic effectiveness} by incorporating chain-of-thought reasoning, allowing developers to explore the inner workings and decision-making processes of AI agents. This fosters a deeper understanding of how strategies are generated and refined, as LLMs can explain the logic behind optimal moves and provide insights into trial-and-error learning.
\item {\it Facilitates collaborative and scalable development} by utilizing LLM functions to generate code and assist in debugging, enabling developers to collaborate on strategy formulation and game design while supporting various approaches. This scalability also encourages team-based improvements and iterative strategy refinement.
\item {\it Provides accessibility for developers with varying expertise} by offering an adaptable framework for hands-on exploration, with LLM-powered tools that simplify experimentation. The tool’s tiered classification, based on Shannon’s taxonomy, coupled with LLM-driven code generation, caters to both novice and advanced programmers. By facilitating dynamic collaboration with LLMs, the system enables efficient real-time generation, optimization, and adaptation of strategies. Ultimately, the goal is to achieve scalable neuralized memoization, enhancing algorithmic efficiency and empowering developers to innovate and refine solutions seamlessly.
\end{itemize}

\begin{figure}[t]
  \centering
  \includegraphics[width=0.48\textwidth]{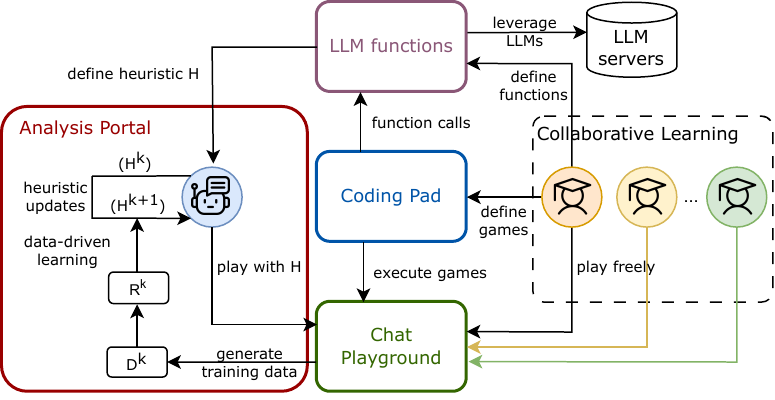}
  \caption{System Flow of Nemobot Games. Students program strategic games, build AI with given heuristics and conduct data-driven AI training.}\label{fig:nemo_sys}
  % \vspace{-6mm}
\end{figure}

\section{Nemobot Game Programming}
\label{sec:nemobot} 

We present Nemobot Games, a programming framework for creating, training, and testing AI game agents on platforms like Facebook Messenger and Telegram. In this section, we provide an architectural overview of the system, detail the interactive training workflow, and describe the tiered curriculum designed to support developers of varying expertise.

\begin{figure*}[h]
  \centering
  \includegraphics[width=\textwidth]{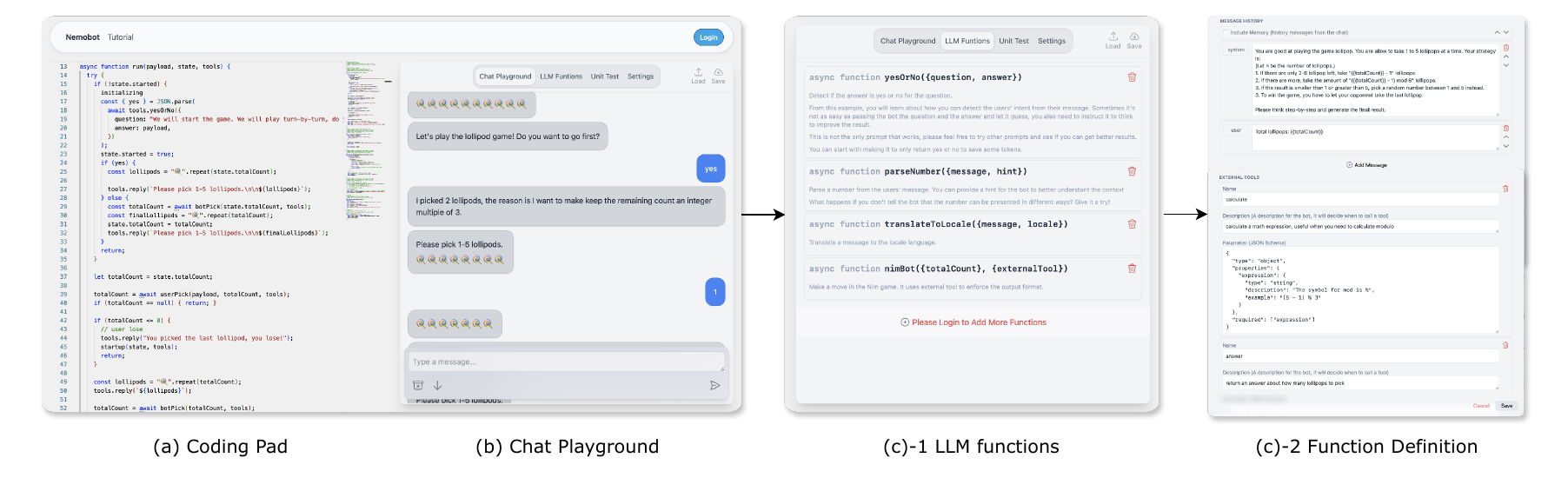}
  \caption{UI of Nemobot Games. (a)\&(b) Programs written on the coding pad are synchronously executed and rendered on the chat playground. (c) LLM functions are defined and called by the main gaming flow.}\label{nemo_ui}
  % \vspace{-2mm}
\end{figure*}

\subsection{Overview of Nemobot Games} \label{sys}
Nemobot provides a dynamic and intuitive environment where developers can design and train AI agents for strategic games, gaining insights into the underlying mechanisms of AI. As illustrated in Figure~\ref{fig:nemo_sys}, the platform comprises a \textbf{coding pad} for programming both the game logic and the AI's reasoning heuristics, a \textbf{chat playground} for interacting with the AI to generate and test data, and an \textbf{analysis portal} for evaluating AI performance and refining heuristics.

The process begins with users designing a two-player strategic game in which players alternate turns, taking actions and accruing rewards until the game concludes. A NodeJS-based template is provided to streamline this process, featuring customizable functions to define the initial state $S_0$, the transition function $f(S_{i+1}|S_{i},a_i)$—where $a_i$ represents the player action at state $S_i$—and the termination state $S_e$. Users then program their AI agent by defining a \textbf{heuristic} $H^0(a_i|S_i)$, which dictates the agent's decision-making in various scenarios.

Nemobot also empowers users to incorporate advanced algorithm design by leveraging LLM functions. Users can bypass complex programming challenges by specifying high-level logic in natural language such as descriptions and examples within the platform. These LLM functions, as shown in Figure~\ref{nemo_ui} part (c), are seamlessly integrated into the main game flow, leveraging LLMs' natural language understanding and code generation capabilities. This feature ensures accessibility for developers with diverse programming backgrounds while maintaining a robust programming paradigm for AI game development.

The coding pad enables users to execute their programs as chatbot-driven games, instantly deployable in the chat playground. This integration facilitates real-time experimentation without switching platforms. Game data is subsequently analyzed and visualized in the analysis portal, offering insights into the AI agent's performance and identifying areas for heuristic refinement. This iterative cycle of programming, testing, and analyzing establishes a robust foundation for data-driven AI training. Additionally, both the programs and training data are shareable, fostering a collaborative development environment at scale.

To address potential concerns associated with Generative AI tools, Nemobot employs a gated interface between users and LLMs. Instead of direct interactions, users define LLM logic within the platform, with Nemobot transmitting prompts using hyper-commands to ensure logical and appropriate responses. This design mitigates over-reliance on Generative AI while maintaining a secure and structured environment for AI programming.

\subsection{System Architecture and Implementation}

The Nemobot system architecture is designed for modularity, low latency, and scalability. The system comprises three core layers:

\begin{itemize}
    \item \textbf{Model-Agnostic AI Integration:} The platform is architected to be independent of any specific Large Language Model (LLM). It operates on a ``Bring Your Own Key'' (BYOK) principle, allowing users to connect via API to various models (e.g., GPT-4, GPT-3.5) depending on their performance requirements and budget. This ensures the framework remains adaptable as model capabilities evolve.
    
    \item \textbf{State Management and Latency:} To ensure a responsive gameplay experience, the game logic and state management are handled locally within the execution environment. During active gameplay, game states are stored in-memory to minimize latency. The LLM is queried only when the user explicitly triggers an ``LLM function'' for strategy construction or reasoning, ensuring that the core game loop remains fast and is not bottlenecked by inference latency.
    
    \item \textbf{Data Persistence and Crowdsourcing:} While individual game states are transient, performance metrics and training outcomes are batched and asynchronously synchronized with a cloud-based database. This allows for the aggregation of crowdsourced data (e.g., win/loss ratios, successful heuristics) across the user base without disrupting real-time interactions, facilitating the collaborative learning described in Section \ref{sec:llmgame}.
\end{itemize}

\subsection{Interactive AI Training}
The core workflow of an AI agent involves perception, processing, and decision-making based on a defined strategy. This strategy can either be pre-programmed with deterministic rules or learned iteratively through methods like reinforcement learning. To support the development of game-playing AI agents, Nemobot provides an interactive framework where users can build and train AI systems from scratch, as outlined in Alg.~\ref{algo_1}.

\begin{algorithm}
\caption{Interactive AI Training with Nemobot Games}\label{algo_1}
\begin{algorithmic}[1]
% \Require $\Sigma_i\eta'_i = 1, \theta_i$ are standard basis vectors
% \Ensure $\tau_H > 0, \tau_{\eta'} > 0, \alpha^n > 0$
\State $H^0 \gets \texttt{initial heuristic},k \gets 1, R \gets \emptyset$
\While{\texttt{true}} 
\For{$i$ \texttt{in} $[1, \texttt{count}(D^k)]$}
\State $R.\texttt{add}(R^k_i \gets \texttt{reward}(D^k_i))$
\EndFor
\If{$\texttt{loss}(R) \leq \tau$}
\State \textbf{break}
\EndIf
\State $H^k \gets \texttt{update}(H^{k-1}, R)$
\State $k \gets (k+1)$
\State {record $<H^k, R>$}
\State $R \gets \emptyset$
\EndWhile \\
\Return $H^k$
\end{algorithmic}
\end{algorithm}
% \vspace{-2mm}

Alg.~\ref{algo_1} illustrates an iterative process where the AI's decision-making heuristic is refined to optimize its performance. During each iteration, the heuristic is evaluated against a dataset $D$ of test cases, yielding rewards $R_i$ that guide adjustments to improve subsequent iterations. This loop continues until the AI meets a predefined performance threshold.

This training paradigm is adaptable to a variety of use cases. For example, in automated machine learning workflows, $\textsf{update()}$ might correspond to gradient-based optimization, while in rule-based AI systems, it could represent manual adjustments to the decision rules. In scenarios where the heuristic is static, the training loop is bypassed entirely. Nemobot Games supports the most general case, enabling end-to-end training for learning-based AI agents while allowing flexibility for specialized implementations.

Users begin by defining an initial heuristic $H^0$, which specifies the AI's actions in a given state and the corresponding rewards. For instance, in the Nim game, $H^0$ might define the AI's move as $n_{AI} = \max(r, 3 - n_{user})$, where $r$ is the remaining count of items and $n_{user}$ is the opponent's last action. Nemobot simplifies this process by leveraging LLM-powered code generation, allowing users to describe heuristics in pseudo-code or natural language rather than executable code. This abstraction lowers the barrier to entry, enabling users to focus on logical reasoning rather than syntax.

Once the heuristic is defined, users can test it in the Chat Playground, where dynamic gameplay and data generation occur simultaneously. The AI is deployed in a chatbot interface, starting from an initial state $s_0$, where users interact with the agent, observing its actions $a_i$ in response to states $s_i$. Final rewards $R^k$ are recorded for analysis. To accelerate testing, Nemobot Games supports automated gameplay, either through random action generation, e.g., simulating ``training by crowdsourcing" or through heuristic-based responses, akin to ``training by another AI."

Heuristic updates are guided by performance analysis. Users refine $H$ iteratively, leveraging LLM functionality to adjust strategies or define new actions. Heuristics can range from precise formulas for deterministic actions to abstract rules representing high-level strategies. Additionally, Nemobot Games supports the incorporation of past training data into the AI’s state representation, enabling more sophisticated decision-making and aligning with modern self-learning methodologies.
% \begin{figure}[h]
%   \centering
%   \includegraphics[width=0.48\textwidth]{samples/images/crowdsourcing.pdf}
%   \caption{Training Results of AI by Crowdsourcing 98 Students. The AI learns to master the Nim Game at Round 43, faster than in normal cases with a single student.}
%   \vspace{-6mm}
% \end{figure}\label{crowdsourcing_results}

% \subsection{Collaborative Learning} 
% \label{collaborative_learning}

\subsection{Tiered Curriculum}
\label{curriculum}
Nemobot Games supports developers with varying expertise by offering structured pathways for crafting AI agents, inspired by Claude E. Shannon's classifications of game-playing machines \cite{shannon1953computers}: dictionary-type machines, complete analysis machines, approximate principle machines, and learning machines. These pathways guide developers from fundamental heuristic design to advanced learning-based AI implementations:
\begin{itemize}
\item Foundational Development: Developers begin by constructing dictionary-type AI agents for predefined games, defining actions $a_i$ for each state $s_i$. This provides insight into how decision-making heuristics influence gameplay outcomes.
\item Intermediate Rule-Based Systems: Using NodeJS templates, developers create custom games and build rule-based AI that follows decision algorithms $f(a_i|s_i)$. Heuristics are iteratively refined based on testing data, introducing the principles of optimization.
\item Advanced Data-Driven AI: Developers employ LLM functions to design complex, adaptive AI systems. By leveraging automated testing methods and crowdsourced training data, they transition to building learning-based AI capable of self-improvement.
\end{itemize}

This structured progression focuses on heuristic-driven decision-making as the foundation and gradually incorporates data-driven methodologies. By simplifying initial development and emphasizing iterative enhancement, Nemobot Games ensures a smooth transition to advanced AI training techniques. Through agentic engineering  where autonomous LLM agents plan, execute, and refine learning strategies, the system can dynamically adapt to individual learner trajectories, selecting appropriate tasks and feedback mechanisms without constant human oversight. 

\section{Evaluation of Nemobot Games}
\label{sec:evaluation}

Nemobot Games uses programmable prompts as a modular, adaptive game-playing agent that adjusts strategies or modes based on user input, game context, or learning goals. For example, in a learning game, Nemobot Games may explain its decision-making process, while in a heuristic game, it synthesizes crowd-sourced strategies for optimal gameplay. Crowd-sourced prompts are categorized by use case (e.g., decision-making, strategy explanation, error correction) and integrated into Nemobot Games as modular components for easy execution and testing. Users can experiment, refine, and contribute to the prompt repository, creating an evolving library that enhances Nemobot Games’s effectiveness and versatility. The web-based programming platform is available for demonstration at: {\tt \url{https://nemobot-neue-experiment.vercel.app}}.

Nemobot Games was evaluated in undergraduate courses at City University of Hong Kong (2021) with 251 students, Nanyang Technological University (2023–2025) with over 80 students, and a remote internship program with around 30 undergraduates from Princeton University (2021–2022). Students initially utilized the OpenAI Playground \cite{WebGPT} and Codex \cite{chen2021evaluating} for prompt engineering, transitioning to ChatGPT \cite{openai_2023} and the OpenAI API in late 2022. Over ten weeks, students learned to program strategic games in Nemobot Games, which can be deployed on platforms like Facebook Messenger and Telegram. Details of some student projects are provided in Table \ref{table:games}. Below, we describe the programming of the strategic games in Nemobot Games, also summarized in Table \ref{table:games}.

\begin{table*}[tb]
    \centering
    \renewcommand{\arraystretch}{1.5}
        \centering
        \newcolumntype{P}[1]{>{\arraybackslash\centering}p{#1}}
        \caption{Nemobot Games, powered by OpenAI Codex and Application Programming Interface (API) between 2021 and 2023, categorized according to Shannon's game taxonomy. These examples demonstrate a versatile range of game programs. The Nemobot Games can be accessed on (\url{https://nemobot-neue-experiment.vercel.app/playground/18}).}
        \label{table:games}
        \begin{tabular}{P{.355\textwidth}P{.355\textwidth}P{.08\textwidth}P{.11\textwidth}}
            \toprule
            \textbf{Game Description} & \textbf{LLM Agent Enhancement} & \textbf{Category} & \textbf{Link} \tabularnewline 
             \midrule
            %To run the Tic Tac Toe game, type the command: \texttt{tic\_tac\_toe hkalexling}.
            % & LLM function to query moves & 1 & \href{https://nemobot-neue-experiment.vercel.app}{Nemobot Game}
             \tabularnewline
            To run the classic Nim game against a human player.
             & LLM function to query moves and optimal strategy algorithm by code generation & 2 & \href{https://nemobot-neue-experiment.vercel.app/playground/18}{Nemobot Games} \tabularnewline
            To run the classic Nim game against LLM agents, press: \texttt{Start Competition}.
             & LLM function to query moves and collect human-AI gameplay data in a leaderboard & 4 & \href{https://nemobot-neue-experiment.vercel.app/playground/18}{Nemobot Games}  \tabularnewline
            Physics learning quiz game challenges players on Newton's three laws.
             & Offer contextual explanations, personalized feedback, hints, and motivational dialogue.
             & 1 & \href{https://github.com/spienaar13/NewtonMechanics}{Newton’s Laws}\tabularnewline
            Single-player RPG where players master math topics by selecting challenges.
             & Use crowdsourcing to offer personalized explanations and answers questions.
             & 3 & \href{https://github.com/glu11/NautilusSummer}{Math Video Recommender}\tabularnewline
            Audio recommendation system game leveraging crowdsourced audio data.
             & NLP search by interpreting user feedback and suggesting improvements.
             & 4 & \href{https://github.com/nautilusinternship/audio-tagging}{Audio Tagging}\tabularnewline
             Web-based chatbot coding to provide peer feedback.
             & Provide coding hints and simulates AI chatbot.
             & 4 & \href{https://github.com/henryherrington/nemobot-labs}{Code Mentor}  \tabularnewline
            Logic-playing game with dynamic dialogue trees.
             & Analyze player inputs and generate logic lessons.
             & 3  & \href{https://github.com/briantieu/NemoBot-Logic-Design}{Logic Playing Game} \tabularnewline \bottomrule
        \end{tabular}
\end{table*}

%\subsection{LLM Gaming Agents}
\subsection{Dictionary-Based Game: Tic Tac Toe}
 Tic Tac Toe is an ideal example of Shannon's first category of game-playing machines, using dictionary-based systems that store precomputed solutions for all possible game states. With a manageable state space of 19,683 configurations, each state is paired with an optimal move. This deterministic approach ensures a well-defined optimal response for every game state, such as maximizing the chances of winning or forcing a draw. While effective, this method can be memory-intensive and complex to implement in rule-based AI games.
 
To address the limitations, we employ the LLM gaming agent, which replicates the behavior of a dictionary-based system without the need to explicitly store an extensive collection of state-action pairs or create complex nested conditional loops. The user interacts with Nemobot Games by making moves, and the LLM responds as the opposing player, maintaining the deterministic characteristics of a dictionary-based system. To facilitate interaction, the $3\times 3$ game board is represented with cells labeled numerically from $0$ to $8$, corresponding to the positions on the board. Users input their moves by specifying the cell where they want to place their mark (e.g., ``$0$'' for the top-left corner or ``$4$'' for the center). To communicate with the LLM agent, the game state is serialized into a structured textual format that captures the positions of X, O, and empty cells. This structured input effectively describes the current state of the game in a manner that the agent can process and respond to. For instance, assuming the LLM agent is already aware of its role and the rules embedded within the configuration of Nemobot Games, the prompt for a given state could be written as below.
% \vspace{2pt}
% \noindent\setlength{\fboxsep}{7pt}\fbox{\begin{minipage}{.94\linewidth}
% \textbf{Example Prompt:}\\
% Current Configuration: Cell 0 (Top-Left) is occupied by 'X'. Cell 4 (Center) is occupied by 'O'. All remaining cells are empty.\\
% Objective: Analyze the board state and execute the optimal move for Player 'O' to prevent a loss or secure a win.
% \end{minipage}}
% \vspace{2pt}

\begin{promptbox}{Example Prompt: Tic Tac Toe Strategy}
    \textbf{Current Configuration:} Cell 0 (Top-Left) is occupied by 'X'. Cell 4 (Center) is occupied by 'O'. All remaining cells are empty.\\
    \textbf{Objective:} Analyze the board state and execute the optimal move for Player 'O' to prevent a loss or secure a win.
\end{promptbox}

Instead of precomputing all possible moves and storing them in a dictionary, the LLM gaming agent uses its pre-trained knowledge and contextual understanding to generate optimal moves dynamically. The backend sends this query to the LLM through an API, and the LLM responds with the most strategic move based on the current configuration. This dynamic querying process eliminates the need for exhaustive storage, effectively compressing the ``dictionary'' into the inference capabilities of the LLM gaming agent. For any given game state, Nemobot Games ensures that the responses of the agent are consistent and deterministic, adhering to the foundational principles of traditional dictionary-based machines. Although the LLM gaming agent dynamically generates responses, these responses align with established optimal strategies for Tic Tac Toe, with its decision-making grounded in its pretrained knowledge. This design ensures that the foundational knowledge required for selecting optimal moves is embedded into the LLM once, allowing it to infer the best moves even for previously unseen positions. This approach significantly reduces memory usage and computational overhead compared to traditional methods that rely on exhaustive searches or evaluating nested conditional loops for each decision. By leveraging the ability of the LLM gaming agent to generalize and interpret game rules, Nemobot Games could achieve enhanced scalability and adaptability, delivering efficient and accurate gameplay without the need for repeated rule-based evaluations.

% \begin{figure}[tbp]
%     \centering
%     % First image
%     \begin{minipage}[b]{0.496\linewidth}
%         \centering
%         \includegraphics[width=\textwidth]{figure/demo3.png}
%         \label{fig:ttt_1}
%     \end{minipage}
%     \hfill
%     % Second image
%     \begin{minipage}[b]{0.49\linewidth}
%         \centering
%         \includegraphics[width=\textwidth]{figure/demo4.png}
%         \label{fig:ttt_2}
%     \end{minipage}
%     \caption{The chatbot interface of Tic Tac Toe.}
%     \label{fig:ttt}
% \end{figure}

\subsubsection{Machines using Rigorous Mathematical Formulas in Nim Game}
Nim is a classic combinatorial game and an example of Shannon's second category of game-playing machines, which use mathematical formulas. Its optimal strategies are based on the Nim-sum, determining winning or losing positions. In Nemobot Games, the player forced to take the last object loses. The AI calculates the Nim-sum at each turn, using the formula to determine the best move: a nonzero Nim-sum indicates a winning position, while a zero Nim-sum signals a losing position. The LLM translates the Nim-sum calculations into natural language, offering insights into the optimal moves and making the game more educational and accessible to users.

The game logic serializes the current state of the board into a structured text format, detailing the number of stones remaining and whose turn it is. The serialized state is then passed to the LLM, which leverages its pretrained knowledge of mathematical solutions for the Nim game to determine the optimal move and contextualize the mathematical strategy. This approach provides the user with a clear explanation of the move, enhancing transparency and ensuring that the reasoning behind the actions took by Nemobot Games is understandable, even in complex scenarios. For example, given a scenario where 8 stones remain, the LLM might receive the following prompt:

% \vspace{2pt}
% \noindent\setlength{\fboxsep}{7pt}\fbox{\begin{minipage}{.94\linewidth} \textbf{Example Prompt:}\\
% Game Status: A single pile remains containing 8 stones.\\
% Constraints: You are permitted to remove 1, 2, or 3 stones. Taking the final stone results in a loss.\\
% Task: Apply the winning mathematical strategy (Nim-sum analysis) to calculate the precise number of stones to remove this turn. Provide a brief rationale for your decision.
% \end{minipage}}
% \vspace{2pt}
\begin{promptbox}{Example Prompt: Nim Game Logic}
    \textbf{Game Status:} A single pile remains containing 8 stones.\\
    \textbf{Constraints:} You are permitted to remove 1, 2, or 3 stones. Taking the final stone results in a loss.\\
    \textbf{Task:} Apply the winning mathematical strategy (Nim-sum analysis) to calculate the precise number of stones to remove this turn. Provide a brief rationale for your decision.
\end{promptbox}

% \begin{figure}[tbp]
%     \centering
%     % First image
%     \begin{minipage}[b]{0.496\linewidth}
%         \centering
%         \includegraphics[width=\textwidth]{figure/demo7.png}
%         \label{fig:n1}
%     \end{minipage}
%     \hfill
%     % Second image
%     \begin{minipage}[b]{0.49\linewidth}
%         \centering
%         \includegraphics[width=\textwidth]{figure/demo8.png}
%         \label{fig:n2}
%     \end{minipage}
%     \caption{The chatbot interface of the Nim game.}
%     \label{fig:nim}
% \end{figure}

\subsubsection{Machines employing General Principles of Play: Mancala Game}
Mancala, a two-player strategy game, exemplifies Shannon's third category of game-playing machines, which rely on general principles of play. Its complex decision space makes precomputing optimal moves impractical, requiring heuristic strategies and game trees to guide decisions. However, this approach is resource-intensive due to the expanding decision space. Nemobot Games addresses these challenges by using LLM functions to dynamically query advice, enhancing game tree strategies without replacing them. The LLM analyzes the current game state to retrieve relevant strategies, reducing computational and memory demands while retaining the game tree as the decision-making foundation.

The game logic begins by converting the current board state into structured text, detailing the number of seeds in each pit, the store values, and the active player. The Mancala board is labeled numerically: pits $0$ to $5$ for the bottom row (left to right) and $7$ to $12$ for the top row (right to left), with stores at $6$ (bottom row) and $13$ (top row). Users select their moves by specifying a pit (e.g., “$0$” for the leftmost pit in the bottom row). For instance, if the bottom-row player has 4 seeds in pit $0$, 1 seed in pit $1$, and 3 seeds in pit $2$, while the top-row player has 2 seeds in pit $7$, 0 seeds in pit $8$, and 4 seeds in pit $9$, with the user’s store holding 8 seeds and the AI’s store holding 6, the serialized prompt provided to the LLM would describe this configuration to generate the next move recommendation.

% \vspace{2pt}
% \noindent\setlength{\fboxsep}{7pt}\fbox{\begin{minipage}{.94\linewidth} \textbf{Example Prompt:}\\
% State Representation:\\
% \textit{Opponent (Bottom Row):} Pit 0: 4 seeds, Pit 1: 1 seed, Pit 2: 3 seeds. Store: 8.\\
% \textit{Agent (Top Row):} Pit 7: 2 seeds, Pit 8: 0 seeds, Pit 9: 4 seeds. Store: 6.\\
% Instruction: Utilize game-tree analysis to determine the move that yields the highest strategic advantage from this position.
% \end{minipage}}
% \vspace{2pt}
\begin{promptbox}{Example Prompt: Mancala Game Tree Analysis}
    \textbf{State Representation:}\\
    \textit{Opponent (Bottom Row):} Pit 0: 4 seeds, Pit 1: 1 seed, Pit 2: 3 seeds. Store: 8.\\
    \textit{Agent (Top Row):} Pit 7: 2 seeds, Pit 8: 0 seeds, Pit 9: 4 seeds. Store: 6.\\
    \textbf{Instruction:} Utilize game-tree analysis to determine the move that yields the highest strategic advantage from this position.
\end{promptbox}

Nemobot Games leverages LLM functions to query optimal paths in the precomputed game tree, retrieving strategies dynamically based on the current game state. For example, in Mancala, if the bottom-row player has 3 seeds in pit 0, 2 in pit 1, and 4 in pit 2, while the top-row player has 1 seed in pit 7, 0 in pit 8, and 5 in pit 9, the LLM identifies the best move, such as playing pit 2 to gain an extra turn. Programmers can obtain different choices of moves that LLM generalizes from similar scenarios, differing from the traditional approach of running the minimax algorithm with the game tree \cite{shannon1950chess}.

\begin{figure}[tbp]
    \centering
    % First image
    \begin{minipage}[b]{0.496\linewidth}
        \centering
        \includegraphics[width=\textwidth]{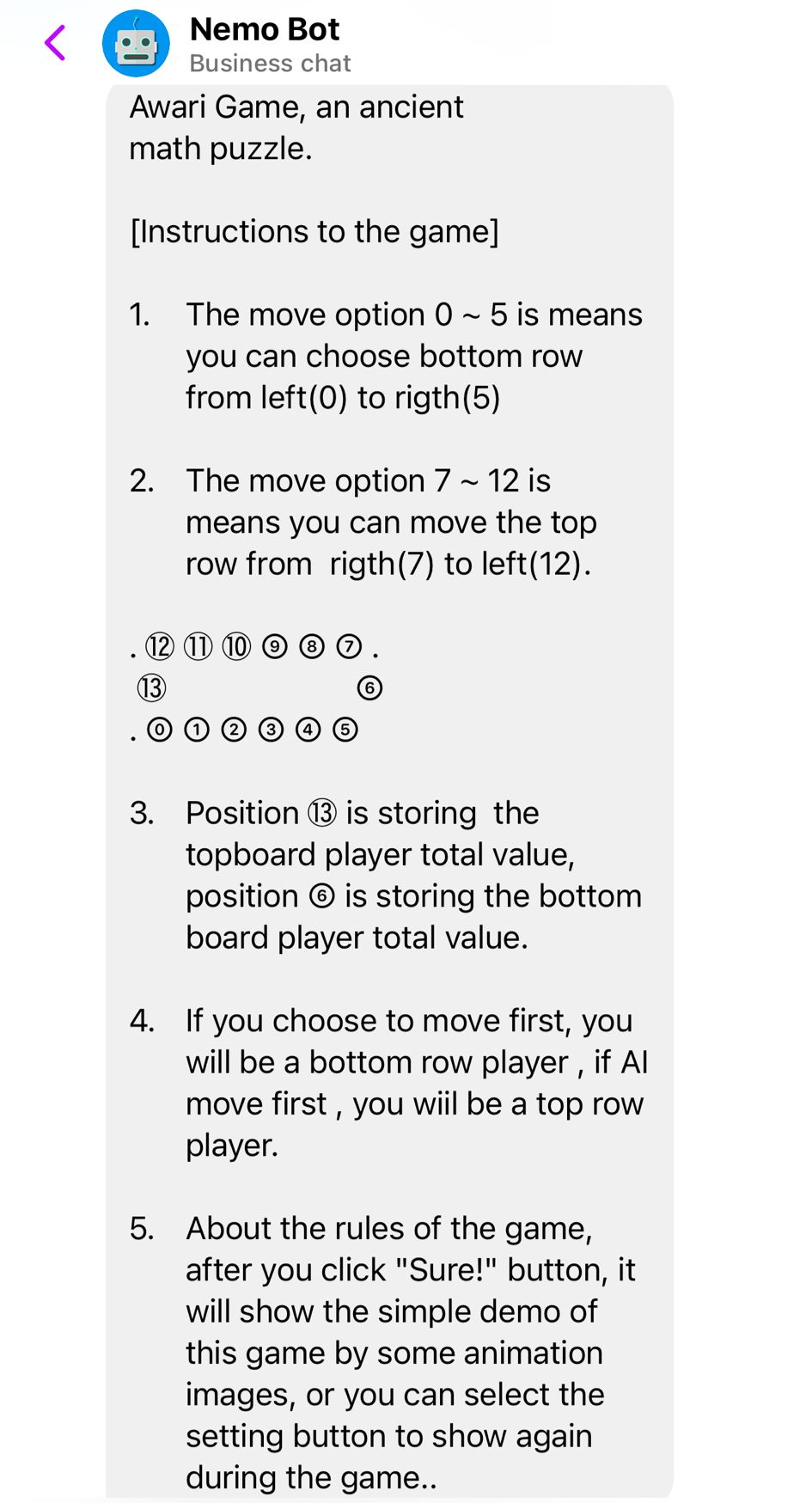}
        \label{fig:a1}
    \end{minipage}
    \hfill
    % Second image
    \begin{minipage}[b]{0.49\linewidth}
        \centering
        \includegraphics[width=\textwidth]{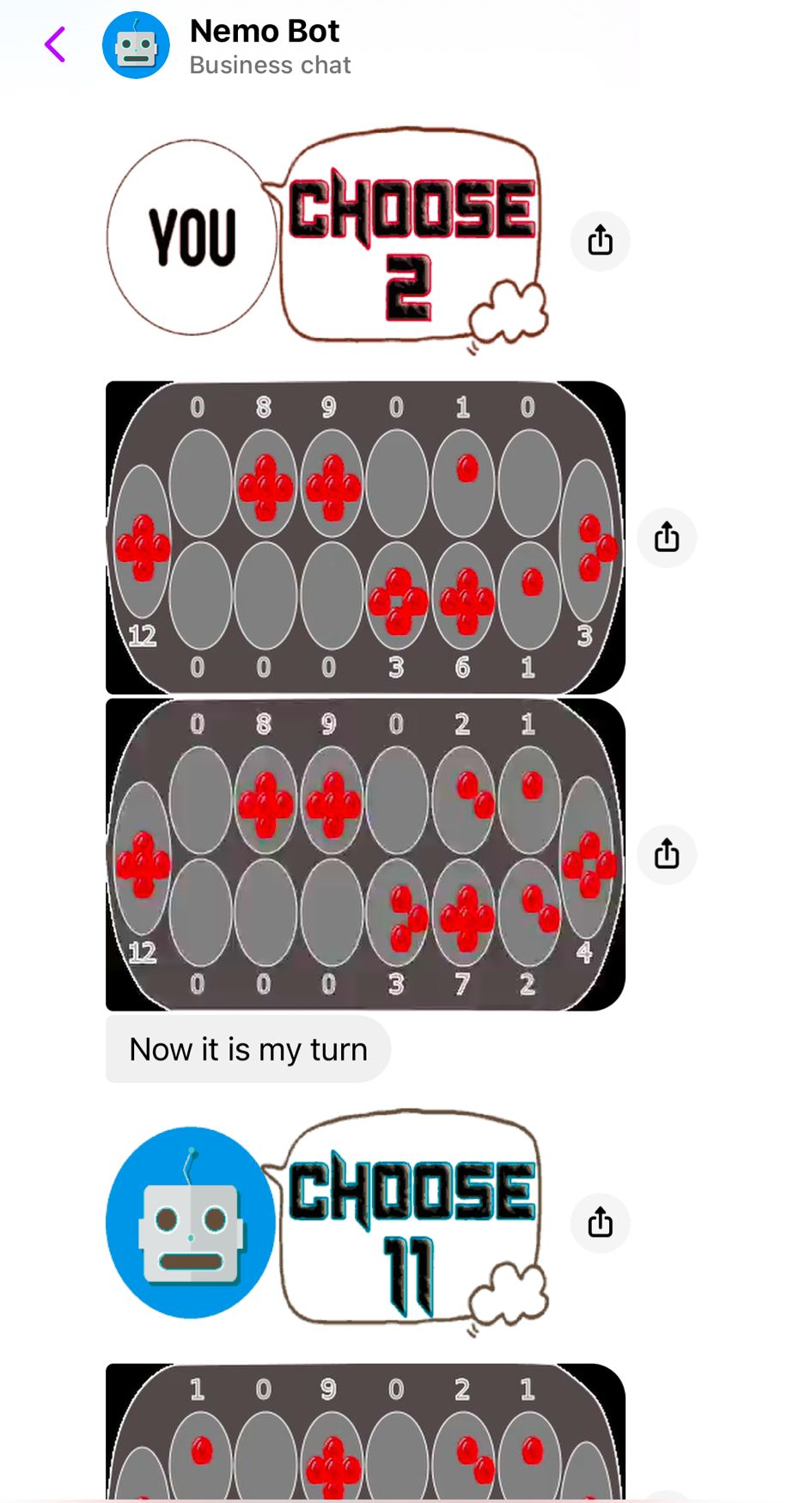}
        \label{fig:a2}
    \end{minipage}
    \caption{The chatbot interface of the Mancala game.}
    \label{fig:awari}
\end{figure}

\subsubsection{Learning Machines by Reinforcement Learning}
Shannon's fourth category, learning machines, improves strategies through experience, using trial-and-error, reinforcement learning, and self-improvement. In games like Nim and Mancala, these systems go beyond predefined rules, leveraging adaptive algorithms and gameplay data. Effective simulations are crucial for gathering high-quality training data, with insights on the number of trials needed and ensuring data accuracy. Fine-tuning LLM functions with this data enhances decision-making, making the AI more efficient and adaptive.

In the case of Nim, a web-based browser simulation is developed to generate extensive gameplay data. The simulation allows players to interact with various initial setups, such as different pile sizes and rules for the number of stones that can be removed per turn. Using this platform, the AI gathers data on optimal strategies and move sequences. The collected gameplay data is then used to fine-tune LLM functions through prompt engineering. For instance, the AI learns to recommend moves based on maximizing its chances of winning, dynamically adapting to game scenarios by referencing its fine-tuned understanding of strategies. A web-based simulation is available for demonstration at: \href{https://nemobot-neue-experiment.vercel.app/playground/18}{Nim Agentic Competition Simulation}.

\begin{figure}[tbp]
    \centering
    \includegraphics[width=0.8\linewidth]{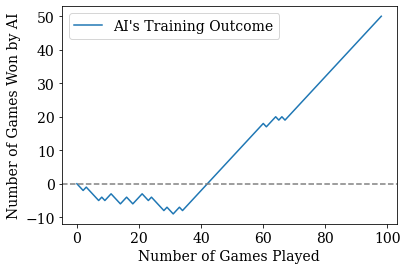}
    \caption{Training process illustrating the number of game trials required for our Nim game agent to master a skillful strategy through a single simulation run of Michie’s algorithm that can be used for LLM function programming. }
    \label{fig:nimgameres}
\end{figure}

\begin{table}[tbp]
    \centering
    \begin{tabular}{|c|ccccccccc|}
    \hline
         N & 5 & 9 & 13 & 17& 21& 10& 16& 19& 22 \\ \hline
         K & 3 & 3 & 3 &  3 &  3&  2&  4&  5&  6 \\  \hline
         L & 5 & 13 & 21 & 29 & 37 &  8 & 39 & 62 & 90 \\ \hline
    \end{tabular}
    \caption{The performance of the Nim Game simulation under various game configurations, where $N$ represents the initial pile size, $K$ is the maximum number of stones that can be taken per turn, and $L$ is the number of rounds required for training. The data is used by the LLM to predict the learning rate of learning by trial-and-error.}
    \label{tab:nimtable}
\end{table}

% \begin{figure}[tbp]
%     \centering
%     \includegraphics[width=0.9\linewidth]{figure/awari_ranking.pdf}
%     \caption{Leaderboard for Mancala Gameplay in Nemobot. Programmers can leverage crowdsourced data to enable self-play and query optimal moves, as shown in the screenshot of the leaderboard with real users competing against the Nemobot Mancala game.}
%     \label{fig:awari_ranking}
% \end{figure}

LLM gaming agents enhance complex games like Mancala through a two-pronged approach: they employ trial-and-error reinforcement learning inspired by Donald Michie's Boxes algorithm \cite{michie1968boxes} to learn optimal moves through experience, while also leveraging crowdsourced move data to refine strategies and accelerate decision-making. In the first prong, simulations employ Donald Michie's Boxes algorithm through self-play. Moves are chosen probabilistically based on the current state's ``box"; winning moves increase the probability of reselection, while losing moves decrease it. Over iterations, this trial-and-error process uncovers optimal strategies across diverse scenarios without requiring expert examples. The second prong involves crowdsourcing, where human players contribute diverse moves by playing against an agent using Shannon's classical minimax algorithm \cite{shannon1950chess}. This exposes the LLM agent to a broader range of tactics and strategies, improving its adaptability to different gameplay styles. Additionally, the LLM agent acts as an instructor, analyzing gameplay, offering real-time feedback on mistakes, and suggesting optimal strategies. This accelerates learning by highlighting suboptimal moves and guiding exploration toward better alternatives, fostering more strategic, long-term decision-making.

To evaluate the performance of the aforementioned approach, we implemented a \textit{crowdsourced leaderboard}, where human players compete against Nemobot. The leaderboard is available at: \href{https://awari.algebragame.app/rating/}{Mancala Leaderboard}. This leaderboard captures data from human-versus-AI matches, where the AI uses the classical minimax algorithm in \cite{shannon1950chess} with three different tree depths corresponding to easy, medium, and hard difficulty levels. The collected data is then used by the LLM to refine its decision-making, either by determining optimal moves or generating self-play algorithms. The leaderboard records player scores and provides valuable insights into the strategies employed by human players. This diverse input ensures that the LLM has access to a wide range of prompts, which can be integrated into its play during the LLM function programming, ultimately enhancing its adaptability and performance.

To facilitate adoption and reproducibility, the Nemobot source code, game implementations, and example prompts are publicly accessible through the project repository. In-platform tutorials and worked examples guide users through the Bring Your Own Key (BYOK) setup, game configuration, and prompt engineering workflows. A live demonstration of the platform is available at: \href{https://nemobot-neue-experiment.vercel.app}{Nemobot Games Demo}.

% \vspace{2pt}
% \noindent\setlength{\fboxsep}{7pt}\fbox{\begin{minipage}{.94\linewidth} \textbf{Example Prompt:}\\
% State Analysis:\\
% \textit{Player (Bottom):} Pits 0 and 1 each contain 2 seeds. Store: 0.\\
% \textit{AI (Top):} Pits 2 and 3 each contain 2 seeds. Store: 0.\\
% Task: It is the AI's turn. Identify the optimal move sequence. Additionally, critique a potential suboptimal move to illustrate strategic errors.\\

% \textbf{Example Response:}\\
% \textit{Recommendation:} Initiate play from Pit 2.\\
% \textit{Strategy:} This action distributes seeds to Pit 3 and Pit 0, landing in the store to trigger a ``free turn'' mechanic. On the subsequent turn, the AI should play from Pit 3, which results in capturing the opponent's seeds in Pit 0.\\
% \textit{Critique:} Starting with Pit 3 immediately is inferior as it forfeits the free turn and fails to capitalize on the capture opportunity. Prioritizing moves that extend the turn sequence is critical for maximizing the score.
% \end{minipage}}
\begin{promptbox}{Example Prompt: Strategic Critique \& Response}
    \textbf{State Analysis:}\\
    \textit{Player (Bottom):} Pits 0 and 1 each contain 2 seeds. Store: 0.\\
    \textit{AI (Top):} Pits 2 and 3 each contain 2 seeds. Store: 0.\\
    \textbf{Task:} It is the AI's turn. Identify the optimal move sequence. Additionally, critique a potential suboptimal move to illustrate strategic errors.
    % \tcblower % This creates a dividing line (optional, or just remove if you prefer continuous text)
    \textbf{Example Response:}\\
    \textit{Recommendation:} Initiate play from Pit 2.\\
    \textit{Strategy:} This action distributes seeds to Pit 3 and Pit 0, landing in the store to trigger a ``free turn'' mechanic. On the subsequent turn, the AI should play from Pit 3, which results in capturing the opponent's seeds in Pit 0.\\
    \textit{Critique:} Starting with Pit 3 immediately is inferior as it forfeits the free turn and fails to capitalize on the capture opportunity. Prioritizing moves that extend the turn sequence is critical for maximizing the score.
\end{promptbox}

\section{Further Discussions}
\label{furtherdiscussions}

\subsection{Implications and Positioning in the Literature}
To situate this work within the broader landscape of LLM-based game AI, it is instructive to compare Nemobot Games with several prominent systems. Voyager~\cite{wang2023voyager} is an open-ended embodied agent that uses GPT-4 to explore and acquire skills autonomously in Minecraft; however, it operates as a fully autonomous agent with no mechanism for human programmers to inspect, modify, or share the strategies it discovers. ChessGPT~\cite{feng2024chessgpt} bridges policy learning and language modeling for chess, yet it is specialized to a single domain and does not expose a programmable interface for educational use. Generative Agents~\cite{park2023generative} simulate believable human behavior through LLM-driven memory and reflection, but their focus is on social simulation rather than strategic game programming. Suspicion-Agent~\cite{guo2024suspicionagent} integrates theory-of-mind reasoning with GPT-4 for imperfect-information games, demonstrating sophisticated multi-agent interactions but without an educational or programming-oriented framework.

In contrast, Nemobot Games occupies a distinct niche: it is a \textit{programmable, educational framework} that exposes LLM functions as transparent, modular components within Shannon's taxonomy. Users do not merely consume AI-generated gameplay; they actively design, test, and refine game-playing strategies through collaborative prompt engineering and crowdsourced feedback. This programmability distinguishes Nemobot from systems that treat LLMs as black-box players and aligns with the broader vision of AI-assisted programming~\cite{wong2023natural,camp,camp_testing} where human oversight and iterative refinement are central to the development process. By grounding its design in Shannon's classical framework and extending it with modern LLM capabilities, Nemobot Games bridges foundational AI theory with contemporary practice, offering both research value and educational utility.

Finally, the algorithmic framework in this paper extends naturally beyond game playing to knowledge tracing and {\it learning-to-learn} in educational AI \cite{cotutor}. In knowledge tracing, a trial-and-error mechanism (e.g., Michie's Boxes algorithm) can model a student's evolving mastery of concepts: each knowledge component becomes a ``box," the student's response (correct or incorrect) serves as reward or punishment, and the probability of selecting a particular strategy updates accordingly--enabling real-time prediction of future performance without requiring expert-labeled skill models. Crowdsourcing, meanwhile, aligns with learning-to-learn by aggregating solution strategies from diverse learners, allowing an intelligent tutoring system to meta-learn which pedagogical approaches work best for which student types. The trained AI agent acting as instructor within a classroom flipping model can become a meta-learning advisor that analyzes a student's error patterns, generates personalized practice problems, and suggests alternative solution pathways \cite{classroomflipping}. This ensures that the complexity of tasks scales effectively with the learner's proficiency. Together, these mechanisms can transform a strategic game-playing AI agent into an adaptive educational system that engages in {\it learning-to-learn}, refining both its predictions and its strategic teaching approaches through interaction with an unknown environment to acquire meta-knowledge about how each individual learns over time.

\subsection{Open Issues and Future Directions}
Future research will expand {\it Nemobot Games} to diverse AI tasks like multimedia generation and programming, enriching AI education. Inspired by Michie’s ideas on discovery as collaboration \cite{michie2000discovery}, future work could develop tools for real-time strategy refinement during gameplay, incorporating adaptive semi-supervised learning with fine-tuning to enable collaboration between three parties--a LLM-enabled copilot, human programmer and a {\it Turing's child machine} in \cite{michie2000discovery} (i.e., a local weaker LLM). Lastly, we list down several open issues remain in developing LLM-powered game-playing agents and collaborative programmable prompt engineering.
\begin{itemize}
    \item LLMs excel in short-term decision-making but face challenges with long-term strategies required for complex games like chess and Mancala, necessitating memory mechanisms or external models. Future research could explore hierarchical systems inspired by Shannon's hierarchy \cite{shannon1953computers}, where learning advances through abstraction. At the highest level, programming copilots could leverage predicate logic for advanced planning, using inductive logic programming to derive predicates from examples and background knowledge \cite{michie1993knowledge}.
    
    \item Another area of exploration involves balancing generalization and specialization. While LLMs are trained on vast datasets to generalize effectively, applying them to specific game domains often requires extensive fine-tuning. Developing adaptive fine-tuning methods that allow for real-time adjustments based on gameplay feedback could enable LLMs to transition seamlessly between different game types or environments.  
    
    \item Additionally, the reliance on human feedback for fine-tuning poses scalability challenges. Although crowdsourced feedback can accelerate learning, ensuring the quality and consistency of this input remains a bottleneck. Future work could explore automated feedback systems or hybrid approaches combining human oversight with AI-generated evaluations.
    
    \item Recent work on adversarial bots \cite{tseng2024robust,wang2023adversarial} shows that, while capable of defeating top Go-playing AI (e.g., AlphaGo \cite{deepmind2019alphastar}), they are trained to exploit hidden vulnerabilities rather than develop well-rounded strategies. A challenge is thus to balance competitive performance with ethical design. Developers must ensure bots engage fairly and transparently, avoid exploiting human weaknesses, and foster challenging yet enjoyable gameplay. Achieving this balance with big data is crucial as AI systems advance, raising important ethical considerations for human-AI interactions in gaming \cite{Donoho2024}.
    
    \item Striking a balance between computational efficiency and reproducibility is a critical challenge, particularly for real-time interactive games that leverage big data. LLMs are highly resource-intensive, making their deployment at scale computationally expensive. To address this, research into lightweight, efficient LLM architectures and hybrid systems is crucial. These systems could offload less complex tasks to smaller, more efficient models while reserving the full capabilities of LLMs for tasks that demand greater sophistication. By achieving this balance, it becomes possible to not only enhance scalability and reduce costs but also ensure reproducibility by enabling consistent benchmarks for performance, resource utilization, and response accuracy across varied applications \cite{Donoho2024}.
\end{itemize}

Addressing these challenges will not only improve AI game systems but also advance broader applications of LLMs in autonomous decision-making and interactive learning.

\section{Conclusion}
\label{sec:conclusion}
In conclusion, this paper introduces a new approach to strategic AI game programming, inspired by Shannon's vision of self-programming automation and Minsky's view of programming as a tool for refining ideas. Central to this is {\it Nemobot Games}, an interactive platform that allows users to create and deploy game agents. By integrating LLMs, {\it Nemobot Games} goes beyond static instructions, enabling AI agents to dynamically adapt to complex gameplay through inherent world models, crowd-sourced insights and iterative learning strategies. This programming paradigm showcases how games in {\it Nemobot Games} surpass rigid algorithms, enabling flexibility and creativity in identifying patterns and evolving strategies in AI games. By autonomously refining their instructions and leveraging crowdsourced data to curate effective prompts, {\it Nemobot Games} can explore complex gameplay challenges for competitive and strategic play. Our approach aligns with Shannon’s vision of self-programming machines, where AI refines strategies through learning. {\it Nemobot Games} bridges AI theory with LLMs, offering educational value and new possibilities for dynamic game design.

\section*{acknowledgments}
This research is supported by the Singapore Ministry of Education Academic Research Fund (MOE-T2EP20224-0009) and NTU startup fund.

\bibliographystyle{IEEEtran}
\bibliography{IEEEabrv,references}

@article{shannon1953computers,
  author = {Shannon, Claude E.},
  title = {Computers and Automata},
  journal = {Proceedings of the IRE},
  volume = {41},
  pages = {1234-1241},
  year = {1953}
}

@article{Shannon1955GamePlaying,
  title={Game playing machines},
  author={Shannon, C. E.},
  journal={Journal of the Franklin Institute},
  volume={260},
  number={6},
  pages={447--453},
  year={1955},
  month={December},
  publisher={Elsevier}
}

@article{shannon1950chess,
  author = {C. E. Shannon},
  title = {Programming a Computer for Playing Chess},
  journal = {Philosophical Magazine},
  volume = {41},
  year = {1950},
  month = {March},
  pages = {256--275}
}

@incollection{minsky1967programming,
  author    = {Marvin Minsky},
  title     = {Why Programming is a Good Medium for Expressing Poorly Understood and Sloppily-Formulated Ideas},
  booktitle = {Design and Planning II: Computers in Design and Communication},
  editor    = {Martin Krampen and Peter Seitz},
  publisher = {Hastings House Publishers},
  address   = {New York},
  year      = {1967},
  pages     = {120--128},
  organization = {Visual Committee Books}
}

@inproceedings{michie1993knowledge,
  title={Knowledge, Learning and Machine Intelligence},
  author={Michie, Donald},
  booktitle={Intelligent Systems},
  pages={63--79},
  year={1993},
  publisher={Springer, Boston, MA}
}

@article{michie1968memofunctions,
  author    = {Donald Michie},
  title     = {'{Memo}' Functions and Machine Learning},
  journal   = {Nature},
  volume    = {218},
  pages     = {19--22},
  year      = {1968},
  doi       = {10.1038/218019a0},
}

@article{barto2019reinforcement,
  author    = {Andrew G. Barto},
  title     = {Reinforcement Learning: Connections, Surprises, Challenges},
  journal   = {AI Magazine},
  volume    = {40},
  number    = {1},
  pages     = {3--15},
  year      = {2019}
}

@article{michie2000discovery,
  author    = {Donald Michie},
  title     = {Discovery as Collaboration: It Takes Two (at Least) to Tango},
  journal   = {Electronic Transactions on Artificial Intelligence},
  volume    = {4},
  number    = {B},
  pages     = {1--19},
  year      = {2000},
}

@book{kasparov2000againstworld,
  author    = {Garry Kasparov and Daniel King},
  title     = {Kasparov Against the World: The Story of the Greatest Online Challenge},
  publisher = {KasparovChess Online, Inc.},
  address   = {New York},
  year      = {2000},
  isbn      = {9780970481306, 0970481306}
}

@article{learningtochat,
  title={Filter, rank, and transfer the knowledge: Learning to chat},
  author={S. Jafarpour and C. J. C. Burges, A. Ritter},
  journal={Advances in Neural Information Processing Systems Workshop on Advances in Ranking},
  volume={10},
  year={2010}
}

@article{christiano2017deep,
  title={Deep Reinforcement Learning from Human Preferences},
  author={Christiano, Paul F and Leike, Jan and Brown, Tom and Martic, Miljan and Legg, Shane and Amodei, Dario},
  journal={Advances in Neural Information Processing Systems},
  volume={30},
  year={2017}
}

@article{chen2021evaluating,
  title={Evaluating large language models trained on code},
  author={Chen, Mark and Tworek, Jerry and Jun, Heewoo and Yuan, Qiming and Pinto, Henrique Ponde de Oliveira and Kaplan, Jared and Edwards, Harri and Burda, Yuri and Joseph, Nicholas and Brockman, Greg and others},
  journal={arXiv preprint arXiv:2107.03374},
  year={2021}
}

@article{WebGPT,
  author    = {Reiichiro Nakano and Jacob Hilton and Suchir Balaji and Jeff Wu and Long Ouyang and Christina Kim and Christopher Hesse and Shantanu Jain and Vineet Kosaraju and William Saunders and Xu Jiang and Karl Cobbe and Tyna Eloundou and Gretchen Krueger and Kevin Button and Matthew Knight and Benjamin Chess and John Schulman},
  title     = {{WebGPT}: Browser-assisted question-answering with human feedback},
  journal   = {CoRR},
  volume    = {abs/2112.09332},
  year      = {2021},
  url       = {https://arxiv.org/abs/2112.09332},
  eprinttype = {arXiv},
  eprint    = {2112.09332}
}

@article{wong2023natural,
  title={Natural Language Generation and Understanding of Big Code for {AI}-Assisted Programming: A Review},
  author={Wong, Man-Fai and Guo, Shangxin and Hang, Ching-Nam and Ho, Siu-Wai and Tan, Chee Wei},
  journal={Entropy},
  volume={25},
  number={6},
  pages={888},
  year={2023},
  publisher={MDPI}
}

@misc{openai_2023, 
    title={CHATGPT: Optimizing Language Models for Dialogue}, 
    url={https://openai.com/blog/chatgpt/}, 
    journal={OpenAI}, 
    publisher={OpenAI}, 
    author={OpenAI}, 
    year={2023}, 
    month={Jan}
}

@misc{friedman2021introducing_url,
  author = {Friedman, Nat},
  title = {Introducing {GitHub Copilot}: Your {AI} Pair Programmer},
  year = {2021},
  howpublished = {\url{https://github.com/features/copilot}},
  note = {Accessed: 2023-07-08}
}

@misc{claude_url, 
     author={{Anthropic Claude}},
     title = {{Claude Code by Anthropic}}, 
     year = {2025},
     howpublished = {\url{https://www.anthropic.com/product/claude-code}},
     note = {Accessed: 2026-03-08}
}

@misc{codex_url, 
 author={{OpenAI}},
 title = {
 {Codex}: A coding agent that helps you build and ship with AI—powered by ChatGPT}, 
 year={2026},
 howpublished = {\url{https://openai.com/codex/}},
 note = {Accessed: 2026-03-08}
}

@article{tancopilot,
  title={{Copilot for Xcode}: Exploring {AI}-Assisted Programming by Prompting Cloud-based Large Language Models},
  author={Tan, Chee Wei and Guo, Shangxin and Wong, Man Fai and Hang, Ching Nam},
  journal={arXiv preprint arXiv:2307.14349},
  year={2023}
}

@inproceedings{WongTan2024,
  author = {Wong, M. F. and Tan, C. W.},
  title = {Aligning Crowd-sourced Human Feedback for Reinforcement Learning on Code Generation by Large Language Model},
  booktitle = {IEEE Transactions on Big Data, to appear},
  pages = {},
  year = {2024}
}

@article{silver2018general,
  title={A general reinforcement learning algorithm that masters chess, shogi, and Go through self-play},
  author={Silver, David and Hubert, Thomas and Schrittwieser, Julian and Antonoglou, Ioannis and Lai, Matthew and Guez, Arthur and Lanctot, Marc and Sifre, Laurent and Kumaran, Dharshan and Graepel, Thore and Lillicrap, Timothy and Simonyan, Karen and Hassabis, Demis},
  journal={Science},
  volume={362},
  number={6419},
  pages={1140--1144},
  year={2018},
  publisher={American Association for the Advancement of Science},
  doi={10.1126/science.aar6404}
}

@article{silver2016mastering,
  title={Mastering the game of Go with deep neural networks and tree search},
  author={Silver, David and Huang, Aja and Maddison, Christopher J and Guez, Arthur and Sifre, Laurent and van den Driessche, George and Schrittwieser, Julian and Antonoglou, Ioannis and Panneershelvam, Veda and Lanctot, Marc and others},
  journal={Nature},
  volume={529},
  pages={484--489},
  year={2016},
  publisher={Springer Nature},
  doi={10.1038/nature16961}
}

@inproceedings{tseng2024robust,
  title={Can Go {AIs} Be Adversarially Robust?},
  author={Tom Tseng and Euan McLean and Kellin Pelrine and Tony T. Wang and Adam Gleave},
  booktitle={Proceedings of the AAAI Conference on Artificial Intelligence},
  volume={39},
  number={26},
  pages={34980},
  year={2025},
  doi={10.1609/aaai.v39i26.34980}
}

@inproceedings{wang2023adversarial,
  title={Adversarial Policies Beat Superhuman Go {AI}s},
  author={Tony T. Wang and Adam Gleave and Tom Tseng and Kellin Pelrine and Nora Belrose and Joseph Miller and Michael D. Dennis and Yawen Duan and Viktor Pogrebniak and Sergey Levine and Stuart Russell},
  booktitle={Proceedings of the 40th International Conference on Machine Learning},
  year={2023},
  pages={202}
}

@article{hu2024survey,
  title={A Survey on Large Language Model-Based Game Agents},
  author={Sihao Hu and Tiansheng Huang and Fatih Ilhan and Selim Tekin and Gaowen Liu and Ramana Kompella and Ling Liu},
  journal={arXiv preprint arXiv:2404.02039},
  year={2024},
  url={https://arxiv.org/abs/2404.02039}
}

@article{bakhtin2022humanlevel,
  title={Human-level play in the game of Diplomacy by combining language models with strategic reasoning},
  author={Meta Fundamental AI Research Diplomacy Team and Anton Bakhtin and Noam Brown and Emily Dinan and Gabriele Farina and Colin Flaherty and Daniel Fried and Andrew Goff and Jonathan Gray and Hengyuan Hu and Athul Paul Jacob and Mojtaba Komeili and Karthik Konath and Minae Kwon and Adam Lerer and Mike Lewis},
  journal={Science},
  year={2022},
  volume={378},
  issue={6624},
  pages={1067--1074}
}

@article{xiao2023llm,
  title={{LLMs} May Not Be Human-Level Players, But They Can Be Testers: Measuring Game Difficulty with {LLM} Agents},
  author={Chang Xiao and Brenda Z. Yang},
  journal={arXiv preprint arXiv:2410.02829},
  year={2024},
  url={https://arxiv.org/abs/2410.02829}
}

@inproceedings{feng2024chessgpt,
  title={{ChessGPT}: Bridging Policy Learning and Language Modeling},
  author={Xidong Feng and Yicheng Luo and Ziyan Wang and Hongrui Tang and Mengyue Yang and Kun Shao and David Mguni and Yali Du and Jun Wang},
  booktitle={Advances in Neural Information Processing Systems},
  volume={36},
  year={2023}
}

@misc{deepmind2019alphastar,
  title={AlphaStar: Mastering the real-time strategy game StarCraft II},
  author={DeepMind},
  year={2019},
  note={\url{https://deepmind.google/discover/blog/alphastar-mastering-the-real-time-strategy-game-starcraft-ii}}
}

@article{achiam2023gpt4,
  title={{GPT-4} Technical Report},
  author={OpenAI and Josh Achiam and Steven Adler and Sandhini Agarwal and Lama Ahmad and Ilge Akkaya and Florencia Leoni Aleman and Diogo Almeida and Janko Altenschmidt and Sam Altman and others},
  journal={arXiv preprint arXiv:2303.08774},
  year={2023},
  url={https://arxiv.org/abs/2303.08774}
}

@article{team2023gemini,
  title={Gemini: A Family of Highly Capable Multimodal Models},
  author={{Gemini Team} and Rohan Anil and Sebastian Borgeaud and Yonghui Wu and Jean-Baptiste Alayrac and Jiahui Yu and Radu Soricut and Johan Schalkwyk and Andrew M. Dai and Anja Hauth and others},
  journal={arXiv preprint arXiv:2312.11805},
  year={2023},
  url={https://arxiv.org/abs/2312.11805}
}

@inproceedings{guo2024suspicionagent,
  title={Suspicion-Agent: Playing Imperfect Information Games with Theory of Mind Aware GPT-4},
  author={Jiaxian Guo and Bo Yang and Paul Yoo and Bill Yuchen Lin and Yusuke Iwasawa and Yutaka Matsuo},
  booktitle={Conference on Logic and Machine Learning},
  year={2024}
}

@inproceedings{yao2022react,
  title={{ReAct}: Synergizing Reasoning and Acting in Language Models},
  author={Shunyu Yao and Jeffrey Zhao and Dian Yu and Nan Du and Izhak Shafran and Karthik Narasimhan and Yuan Cao},
  booktitle={International Conference on Learning Representations},
  year={2023}
}

@inproceedings{wang2022selfconsistency,
  title={Self-Consistency Improves Chain of Thought Reasoning in Language Models},
  author={Xuezhi Wang and Jason Wei and Dale Schuurmans and Quoc V. Le and Ed H. Chi and Sharan Narang and Aakanksha Chowdhery and Denny Zhou},
  booktitle={International Conference on Learning Representations},
  year={2023}
}

@inproceedings{besta2023graphoftoughts,
  title={Graph of Thoughts: Solving Elaborate Problems with Large Language Models},
  author={Maciej Besta and Nils Blach and Ales Kubicek and Robert Gerstenberger and Michal Podstawski and Lukas Gianinazzi and Joanna Gajda and Tomasz Lehmann and Hubert Niewiadomski and Piotr Nyczyk and Torsten Hoefler},
  booktitle={Proceedings of the AAAI Conference on Artificial Intelligence},
  volume={38},
  number={16},
  pages={17682--17690},
  year={2024},
  doi={10.1609/aaai.v38i16.29720}
}

@inproceedings{wu2024spring,
  title={{SPRING}: Studying the Paper and Reasoning to Play Games},
  author={Yue Wu and Shrimai Prabhumoye and So Yeon Min and Yonatan Bisk and Ruslan Salakhutdinov and Amos Azaria and Tom Mitchell and Yuanzhi Li},
  booktitle={Advances in Neural Information Processing Systems},
  volume={36},
  year={2023}
}

@article{Donoho2024,
  author    = {David Donoho},
  title     = {Data Science at the Singularity},
  journal   = {Harvard Data Science Review},
  volume    = {6},
  number    = {1},
  year      = {2024},
}

@article{coledaviegame,
  author    = {Cole, A. J. and Davie, A. J. T.},
  title     = {A game based on the Euclidean algorithm and a winning strategy for it},
  journal   = {Mathematical Gazette},
  volume    = {53},
  number    = {386},
  pages     = {354--357},
  year      = {1969}
}

@article{chainofthoughts,
  author    = {Wei, Jason and Wang, Xuezhi and Schuurmans, Dale and Bosma, Maarten and Ichter, Brian and Xia, Fei and Chi, Ed H. and Le, Quoc and Zhou, Denny},
  title     = {Chain-of-Thought Prompting Elicits Reasoning in Large Language Models},
  journal   = {arXiv preprint arXiv:2201.11903},
  year      = {2022},
  url       = {https://arxiv.org/abs/2201.11903}
}

@article{huang2024agent,
  title={Agent AI Towards a Holistic Intelligence},
  author={Qiuyuan Huang and Naoki Wake and Bidipta Sarkar and Zane Durante and Ran Gong and Rohan Taori and Yusuke Noda and Demetri Terzopoulos and Noboru Kuno and Ade Famoti and Ashley J. Llorens and John Langford and Hoi Vo and Fei-Fei Li and Katsushi Ikeuchi and Jianfeng Gao},
  journal={arXiv preprint arXiv:2403.00833},
  year={2024},
  url={https://arxiv.org/abs/2403.00833},
  note={Accessed: 2024-12-15}
}

@article{michie1968boxes,
  title={BOXES: An experiment in adaptive control},
  author={Michie, Donald and Chambers, Roger A},
  journal={Machine intelligence},
  volume={2},
  number={2},
  pages={137--152},
  year={1968},
  publisher={Citeseer}
}

@article{michie1973machines,
  title={Machines and the theory of intelligence},
  author={Michie, Donald},
  journal={Nature},
  volume={241},
  number={23.02},
  pages={1973},
  year={1973},
  publisher={Nature Publishing Group UK London}
}

@article{wang2023voyager,
  title={Voyager: An Open-Ended Embodied Agent with Large Language Models},
  author={Wang, Guanzhi and Xie, Yuqi and Jiang, Yunfan and Mandlekar, Ajay and Xiao, Chaowei and Zhu, Yuke and Fan, Linxi and Anandkumar, Anima},
  journal={arXiv preprint arXiv:2305.16291},
  year={2023}
}

@inproceedings{park2023generative,
  title={Generative Agents: Interactive Simulacra of Human Behavior},
  author={Park, Joon Sung and O'Brien, Joseph C and Cai, Carrie J and Morris, Meredith Ringel and Liang, Percy and Bernstein, Michael S},
  booktitle={Proceedings of the 36th Annual ACM Symposium on User Interface Software and Technology},
  pages={1--22},
  year={2023}
}

@inproceedings{li2022nemobot,
  title        = {A Chatbot-Server Framework for Scalable Machine Learning Education through Crowdsourced Data},
  author       = {Li, J. and Tan, C. W. and Hang, C. and Qi, X.},
  booktitle    = {Proceedings of the Ninth ACM Conference on Learning @ Scale (L@S '22)},
  year         = {2022},
  publisher    = {ACM},
  address      = {New York, NY, USA}
}

@inproceedings{nemobot2024,
  title={Nemobot: Crafting Strategic Gaming {LLM} Agents for K-12 {AI} Education},
  author={Wang, Yuchen and Guo, Shangxin and Ling, Lin and Tan, Chee Wei},
  booktitle={Proceedings of the Eleventh ACM Conference on Learning@ Scale},
  pages={393--397},
  year={2024}
}

@inproceedings{camp,
  title={Contextual augmented multi-model programming {(CAMP)}: A local-cloud copilot solution},
  author={Wang, Yuchen and Guo, Shangxin and Tan, Chee Wei},
  booktitle={2025 IEEE Conference on Artificial Intelligence (CAI)},
  pages={675--681},
  year={2025},
  organization={IEEE}
}

@article{camp_testing,
  title={From code generation to software testing: {AI} Copilot with context-based {RAG}},
  author={Wang, Yuchen and Guo, Shangxin and Tan, Chee Wei},
  journal={IEEE Software},
  year={2025},
  volume={42},
  number={4},
  pages={34-42},
  publisher={IEEE}
}

@article{cotutor,
  title={Future-Proofing Programmers: Optimal Knowledge Tracing for AI-Assisted Personalized Education},
  author={Wang, Yuchen and Yu, Pei-Duo and Tan, Chee Wei},
  journal={IEEE Signal Processing Magazine},
  volume={43},
  number={1},
  pages={69-82},
  publisher={IEEE},
  year={2026}
}

@article{classroomflipping,
  title={Large language model-driven classroom flipping: Empowering student-centric peer questioning with flipped interaction},
  author={Tan, Chee Wei},
  journal   = {CoRR},
  volume    = {abs/2311.14708},
  year      = {2023},
  url       = {https://arxiv.org/abs/2311.14708},
  eprinttype = {arXiv}
}

@inproceedings{toshniwal2022chess,
  title = {Chess as a Testbed for Language Model State Tracking},
  author = {Toshniwal, Shubham and Wiseman, Sam and Livescu, Karen and Gimpel, Kevin},
  booktitle = {Proceedings of the {AAAI} Conference on Artificial Intelligence},
  volume = {36},
  pages = {11385--11393},
  year = {2022}
}

@inproceedings{li2023emergent,
  title = {Emergent World Representations: Exploring a Sequence Model Trained on a Synthetic Task},
  author = {Li, Kenneth and Hopkins, Aspen K. and Bau, David and Vi{\'e}gas, Fernanda and Pfister, Hanspeter and Wattenberg, Martin},
  booktitle = {International Conference on Learning Representations},
  year = {2023}
}

@inproceedings{vafa2024evaluating,
  title = {Evaluating the World Model Implicit in a Generative Model},
  author = {Vafa, Keyon and Chen, Justin Y. and Rambachan, Ashesh and Kleinberg, Jon and Mullainathan, Sendhil},
  booktitle = {Advances in Neural Information Processing Systems},
  volume = {37},
  year = {2024},
  note = {To appear in {NeurIPS} 2024}
}

@article{phillips2025goal,
  title={Goal-Oriented Interactions in Games Using {LLMs}},
  author={Adon Phillips and Jochen Lang and David Mould},
  journal={IEEE Transactions on Games},
  volume={17},
  number={2},
  pages={510--521},
  year={2025},
  month={June},
  doi={10.1109/TG.2024.3515807}
}

@article{bassanelli2025gamidoc,
  title={{GamiDOC}: The Importance of Designing Gamification in a Proper Way},
  author={Simone Bassanelli and Antonio Bucchiarone and Federica Gini},
  journal={IEEE Transactions on Games},
  volume={17},
  number={1},
  pages={13--31},
  year={2025},
  month={March},
  doi={10.1109/TG.2024.3364061}
}

\begin{IEEEbiography}{Chee Wei Tan} (~\IEEEmembership{Senior Member,~IEEE}) received his M.A. and Ph.D. in Electrical Engineering from Princeton University. He is an Associate Professor at Nanyang Technological University. He conducts research in networks, distributed optimization, and generative AI. Dr. Tan has served as IEEE Distinguished Lecturer of IEEE Communications Society and was selected twice for the U.S. National Academy of Engineering China-America Frontiers of Engineering Symposium.
\end{IEEEbiography}

\begin{IEEEbiography}{Yuchen Wang}
(~\IEEEmembership{Student Member,~IEEE}) is a Ph.D. candidate at Nanyang Technological University and a full-stack software engineer specializing in mobile development. She received her B.E. and M.S. degrees in Computer Science, both with Honours with the Highest Distinction, from the National University of Singapore. Her research mainly focuses on AI-Assisted Programming and Sound and Music Computing.  
\end{IEEEbiography}

\begin{IEEEbiography}{Shangxin Guo}{\,} is a senior software developer specializing in macOS and iOS development since 2015. He received his B.E. degree from Shenzhen University and later obtained an M.S. degree from City University of Hong Kong. Currently, he is primarily focused on developing AI applications, specifically Copilot for Xcode, an AI code assistant.\vspace*{8pt}
\end{IEEEbiography}

\end{document}